\documentclass[a4paper,conference]{IEEEtran}
\usepackage{cite}
\ifCLASSINFOpdf
\usepackage{graphicx}
\graphicspath{{./images/}}
\else
\fi
\usepackage{array}
\ifCLASSOPTIONcompsoc
  \usepackage[caption=false,font=normalsize,labelfont=sf,textfont=sf]{subfig}
\else
  \usepackage[caption=false,font=footnotesize]{subfig}
\fi

\usepackage{float}
\usepackage{placeins}
\usepackage{afterpage}
\begin{document}
\title{On the Impact of Lossy Image and Video Compression on the Performance of Deep Convolutional Neural Network Architectures}

\author{\IEEEauthorblockN{Matt Poyser, Amir Atapour-Abarghouei, Toby P. Breckon}
\IEEEauthorblockA{Department of Computer Science\\
Durham University, UK.}}


%Telephone: (800) 555--1212, Fax: (888) 555--1212}


\maketitle

%%%%%%%%%%%%%%%%%%%%%%%%ABSTRACT%%%%%%%%%%%%%%%%%%%%%%%%%%%%%%%%
%%%%%%%%%%%%%%%%%%%%%%%%%%%%%%%%%%%%%%%%%%%%%%%%%%%%%%%%%%%%%%%%%%%%
\begin{abstract}
	Recent advances in generalized image understanding have seen a surge in the use of deep convolutional neural networks (CNN) across a broad range of image-based detection, classification and prediction tasks. Whilst the reported performance of these approaches is impressive, this study investigates the hitherto unapproached question of the impact of commonplace image and video compression techniques on the performance of such deep learning architectures. Focusing on the JPEG and H.264 (MPEG-4 AVC) as a representative proxy for contemporary lossy image/video compression techniques that are in common use within network-connected image/video devices and infrastructure, we examine the impact on performance across five discrete tasks: human pose estimation, semantic segmentation, object detection, action recognition, and monocular depth estimation. As such, within this study we include a variety of network architectures and domains spanning end-to-end convolution, encoder-decoder, region-based CNN (R-CNN), dual-stream, and generative adversarial networks (GAN). Our results show a non-linear and non-uniform relationship between network performance and the level of lossy compression applied. Notably, performance decreases significantly below a JPEG quality (quantization) level of 15\% and a H.264 Constant Rate Factor (CRF) of 40. However, retraining said architectures on pre-compressed imagery conversely recovers network performance by up to 78.4\% in some cases. Furthermore, there is a correlation between architectures employing an encoder-decoder pipeline and those that demonstrate resilience to lossy image compression. The characteristics of the relationship between input compression to output task performance can be used to inform design decisions within future image/video devices and infrastructure.
\end{abstract}


% \fi
\IEEEpeerreviewmaketitle

%%%%%%%%%%%%%%%%%%%%%%%%INTRODUCTION%%%%%%%%%%%%%%%%%%%%%%%%%%%%%%%%
%%%%%%%%%%%%%%%%%%%%%%%%%%%%%%%%%%%%%%%%%%%%%%%%%%%%%%%%%%%%%%%%%%%%
\section{Introduction}
Image compression is in \textit{de facto} use within environments relying upon efficient image and video transmission and storage such as security surveillance systems within our transportation infrastructure and our daily use of mobile devices. However, the use of the commonplace lossy compression techniques, such as JPEG \cite{JPEG} and MPEG \cite{MPEG} to lower the storage/transmission overheads for such smart cameras leads to reduced image quality that is either noticeable or commonly undetectable to the human observer. With the recent rise of deep convolutional neural networks (CNN \cite{Hinton, nature}) for video analytics across a broad range of image-based detection applications, a primary consideration for classification and prediction tasks is the empirical trade-off between the performance of these approaches and the level of lossy compression that can be afforded within such practical system deployments (for storage/transmission).

This is of particular interest as CNN are themselves known to contain lossy compression architectures - removing redundant image information to facilitate both effective feature extraction and retaining an ability for full or partial image reconstruction from their internals \cite{Hinton, nature}.

Prior work on this topic \cite{AVG, highspeedHAR, faceBitRate, faceDetectionh264} largely focuses on the use of compressed imagery within the train and test cycle of deep neural network development for specific tasks.  However, relatively few studies investigate the impact upon CNN task performance with respect to differing levels of compression applied to the input imagery at inference (deployment) time.

In this paper we investigate whether (a) existing pre-trained CNN models exhibit linear degradation in performance as image quality is impacted by the use of lossy compression and (b) whether training CNN models on such compressed imagery thus improves performance under such conditions. In contrast to prior work topic \cite{AVG, highspeedHAR, faceBitRate, faceDetectionh264}, we investigate these aspects across multiple CNN architectures and domains spanning segmentation (SegNet, \cite{SegNet}), human pose estimation (OpenPose, \cite{OpenPose}), object recognition (R-CNN, \cite{FasterRCNN}), human action recognition (dual-stream, \cite{HAR}), and depth estimation (GAN, \cite{Depth}). Furthermore, we determine within which domains compression is most impactful to performance and thus where image quality is most pertinent to deployable CNN model performance.



%%%%%%%%%%%%%%%%%%%%%%%%Lit REVIEW%%%%%%%%%%%%%%%%%%%%%%%%%%%%%%%%
%%%%%%%%%%%%%%%%%%%%%%%%%%%%%%%%%%%%%%%%%%%%%%%%%%%%%%%%%%%%%%%%%%%%
\section{Prior Work}
Overall, prior work in this area is limited in scope and diversity \cite{AVG, highspeedHAR, faceBitRate, faceDetectionh264}. Dodge et al. \cite{AVG} analyze the performance of now seminal CNN image classification architectures (AlexNet \cite{alexnet}, VGG \cite{vggnet} and InceptionV1 \cite{inception}) performance under JPEG \cite{JPEG} compression and other distortion methods. They find that these architectures are resilient to compression artifacts (performance drops only for JPEG quality $<$ 10) and contrast changes, but under-perform when noise and blur are introduced.

Similarly, Zanjani et al. \cite{retrainJPEG} consider the impact of JPEG 2000 compression \cite{JPEG2000} on CNN, and whether retraining the network on lossy compressed imagery would afford better resultant model performance. They identify similar performance from the retrained model on higher quality images but are able to achieve up to as much as 59\% performance increase on low quality images.

Rather than image compression, Yeo et al. \cite{highspeedHAR} compare different block sizes and group-of-pictures (GOP) sizes within MPEG \cite{MPEG} compression against Human Action Recognition (HAR). They determine that both smaller blocks and smaller groups increase performance. Furthermore, B frames introduce propagation errors in computing block texture, and should be avoided within the compression process. Tom et al. \cite{HARMV} add that there is a near-linear relationship between HAR performance and the number of motion vectors (MV) corrupted within H.264\cite{h264} video data, with performance levelling off when 75\% of MV are corrupted. Klare and Burge \cite{faceBitRate}, however, demonstrate that there is a non-linear relationship between face recognition performance and bit rate within H.264 video data, with sudden performance degradation around 128kbps (CRF). These contrasting results therefore demonstrate the need to investigate compression quality across multiple challenge domains, whose respective model architectures might have different resilience to lossy compression artifacts.

Multiple authors have developed impressive architectures trained on compressed data, indicating both the potential and need for in-depth investigation within the compressed domain. Zhuang and Lai \cite{faceDetectionh264} demonstrate that acceptable face detection performance can be obtained from H.264 video data, while Wang and Chang \cite{facedetectionmpeg} use the DCT coefficients from MPEG compression \cite{MPEG} to directly locate face regions. The same authors even achieve accurate face tracking results in \cite{faceTrack}, still within the compressed video domain. The question is evidently:- by \textit{how much} can data be compressed?

These limited studies open the door only slightly on this very question - \textit{what is generalized impact of compression on varying deep neural network architectures?} Here we consider multiple CNN variants spanning region-based, encoder-decoder and GAN architectures in addition to a wide range of target tasks spanning both discrete and regressive outputs. From our observations, we aim to form generalized conclusions on the hitherto unknown relationship between (lossy) image input to target function outputs within the domain of contemporary CNN approaches. 
% TPB read and corrected up to here

%%%%%%%%%%%%%%%%%%%%%%%%EXPERIMENT%%%%%%%%%%%%%%%%%%%%%%%%%%%%%%%%
%%%%%%%%%%%%%%%%%%%%%%%%%%%%%%%%%%%%%%%%%%%%%%%%%%%%%%%%%%%%%%%%%%%%
\section{Methodology}
To determine how much lossy image compression is viable within CNN architectures before performance is significantly impacted we must study a range of second generation tasks, beyond simple and holistic image classification, requiring more complex CNN output granularity. We examine five CNN architectural variants across five different challenge domains, emulating the dataset and evaluation metrics characterized in their respective originating study in each case as closely as possible. Inference models processing images were tested six times, with a JPEG quality parameter in the set $\{5, 10, 15, 50, 75, 95\}$, while video-based models were tested with H.264 CRF compression parameters in the set $\{23, 25, 30, 40, 50\}$. Each model is then retrained with imagery compressed at each of the five higher levels of lossy compression to determine whether resilience to compression could be improved, and how much compression we can afford before a significant impact on performance is observed. Our methodology for each of our representative challenge domains is outlined in the following sections:- semantic segmentation (Section: \ref{sec:segmentation}), depth estimation (Section: \ref{sec:depth}), object detection (Section: \ref{sec:fasterrcnn}), human pose estimation (Section: \ref{sec:pose}), and human action recognition (Section: \ref{sec:har}).

\subsection{Semantic Segmentation} \label{sec:segmentation}
Pixel-wise Segmantic segmentation involves assigning each pixel in an image (Fig. \ref{fig:segm95}, above) its respective class label (Fig. \ref{fig:segm95}, below). SegNet \cite{SegNet} uses an encoder-decoder neural network architecture followed by a pixel-wise classification layer to approach this challenge.

Implementing SegNet from \cite{pytorchSegNet}, we evaluate global accuracy (percentage of pixels correctly classified), mean class accuracy (mean prediction accuracy over each class), and mean intersection over union (mIoU) against compressed imagery from the Cityscapes dataset \cite{cityscapes}. When retraining the network, we use 33000 epochs, with a batch size of 12, fixed learning rate ($\eta$) of 0.1, and momentum ($\beta$) of 0.9. 

\begin{figure}[!t]
	\centering
	\caption{Results of pre-trained SegNet model \cite{SegNet} on a JPEG image under different compression levels (original RGB image above, computed segmentation map below)}
	\label{fig:segm}
	\setcounter{figure}{0}
	\subfloat[JPEG compression level: 95]{
		\begin{tabular}{c}
			\includegraphics[trim={2.5cm 2.5cm 2.5cm 2.5cm},clip, width=0.8\linewidth]{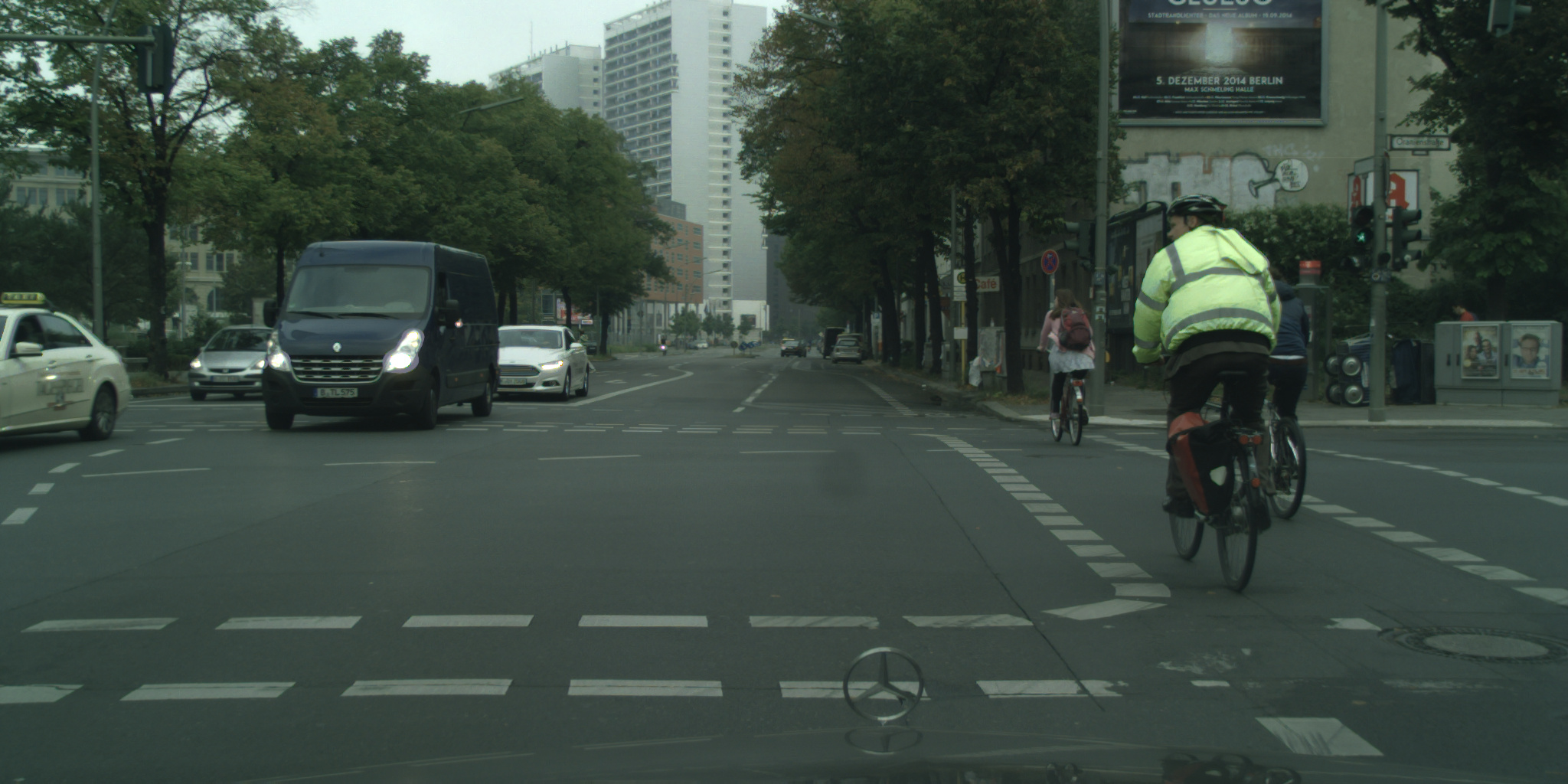}\\
			\includegraphics[trim={1.5cm 1.5cm 1.5cm 1.5cm},clip, width=0.8\linewidth]{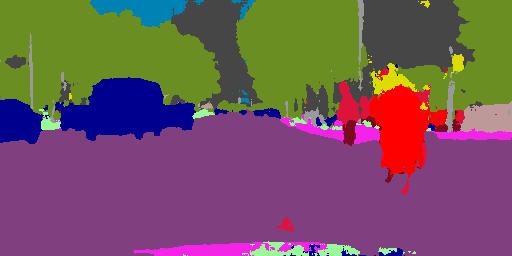}\\
		\end{tabular}
		\label{fig:segm95}}
	\vfil
	\subfloat[JPEG compression level: 15]{
		\begin{tabular}{c}
			\includegraphics[trim={2.5cm 2.5cm 2.5cm 2.5cm},clip, width=0.8\linewidth]{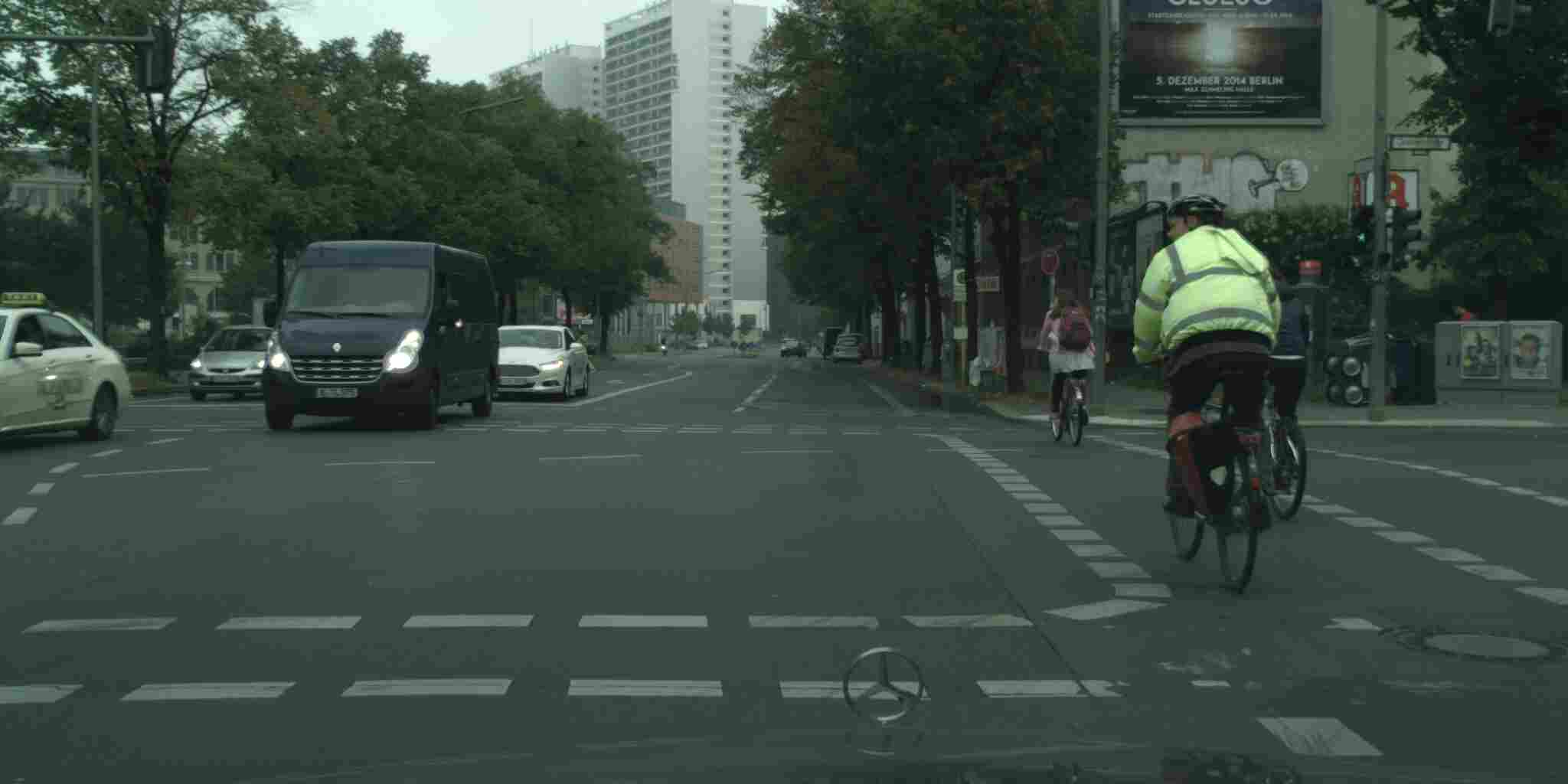}\\
			\includegraphics[trim={1.5cm 1.5cm 1.5cm 1.5cm},clip, width=0.8\linewidth]{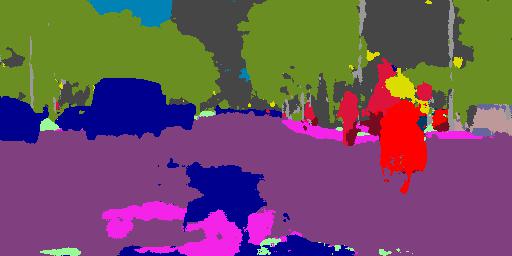}\\
		\end{tabular}
		\label{fig:segm15}}
	\vfil
	\subfloat[JPEG compression level: 10]{
		\begin{tabular}{c}
			\includegraphics[trim={2.5cm 2.5cm 2.5cm 2.5cm},clip, width=0.8\linewidth]{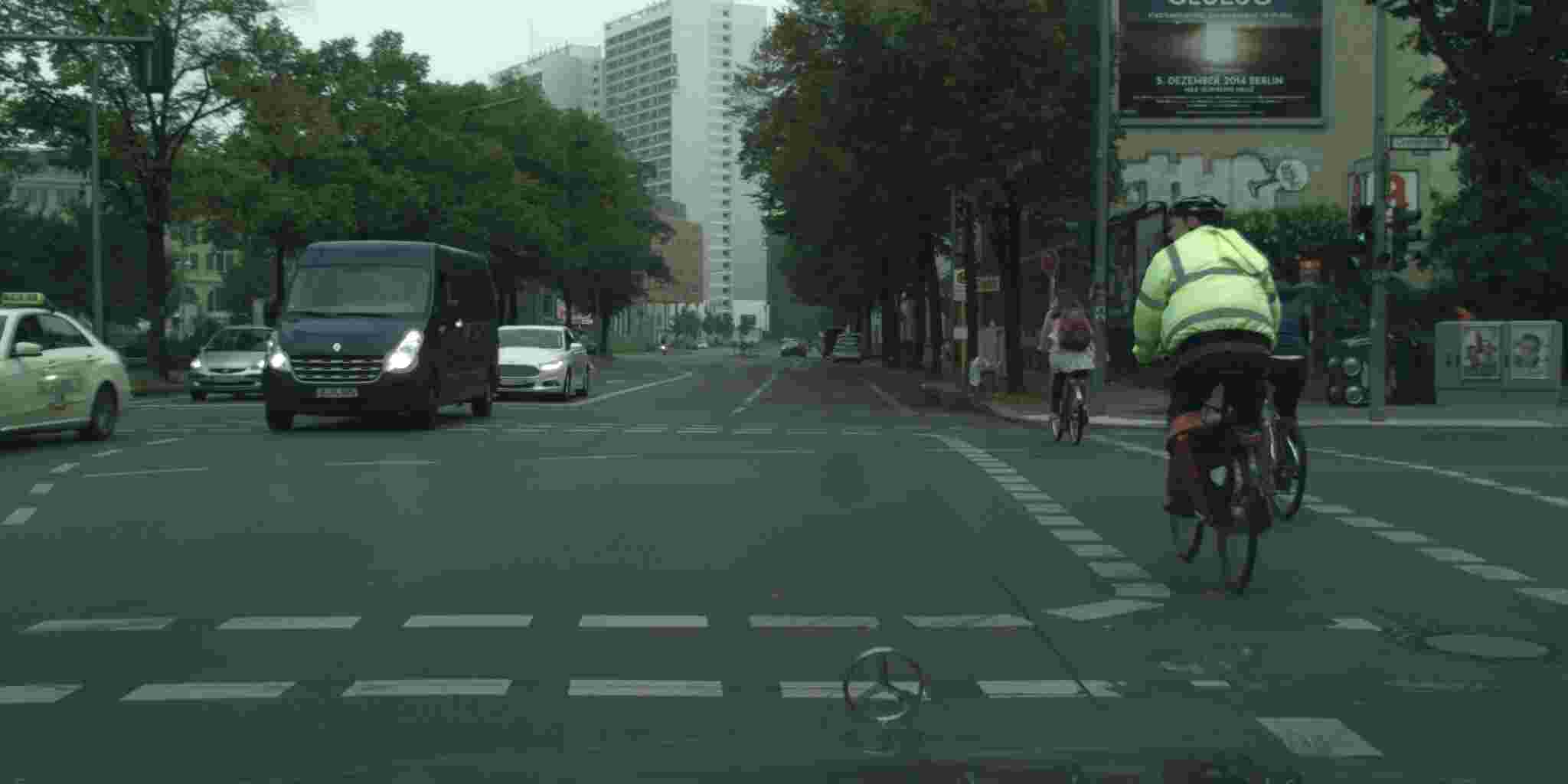}\\
			\includegraphics[trim={1.5cm 1.5cm 1.5cm 1.5cm},clip, width=0.8\linewidth]{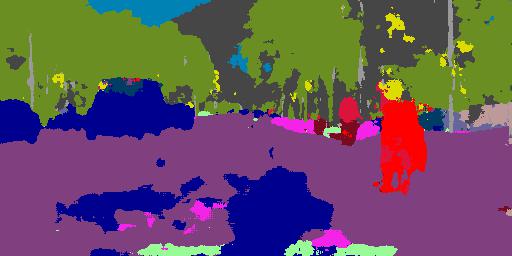}
		\end{tabular}
		\label{fig:segm10}}
	\setcounter{figure}{1}
\end{figure}

\subsection{Depth Estimation} \label{sec:depth}
In order to evaluate GAN architecture performance under compression, we need a task decoupled from reconstructing high quality output, to which compression would be clearly detrimental. One such example is computing the depth map of a scene (Fig. \ref{fig:depth95}, below) from monocular image sequences (Fig. \ref{fig:depth95}, above). 

Using a simplified network from \cite{Depth}, we evaluate RMSE performance of the GAN against the Synthia dataset presented in \cite{synthia}. We employ $\eta=0.0001$ and batch size 10 over 10 epochs.

\begin{figure}[!t]
	\centering
	\caption{Results of pre-trained GAN model on a JPEG image under different compression levels (RGB image above, computed depth map below)}
	\label{fig:depth}
	\setcounter{figure}{1}
	\setlength\extrarowheight{-10pt}
	\subfloat[JPEG compression level: 95]{
		\begin{tabular}{c} 
			\includegraphics[width=0.95\linewidth]{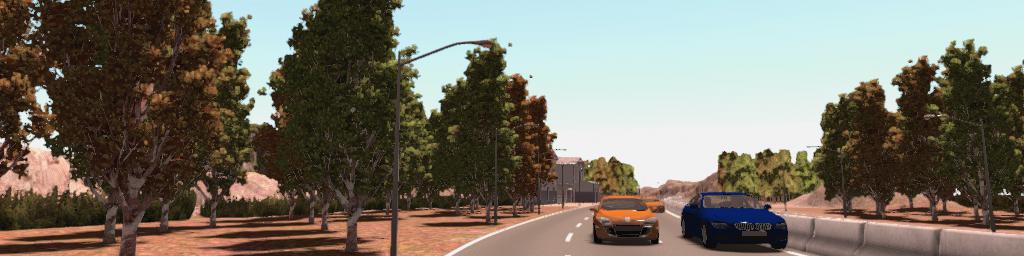}\\
			\includegraphics[width=0.95\linewidth]{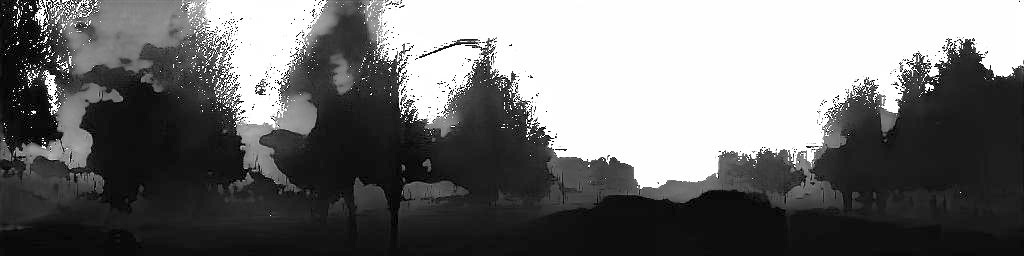}
		\end{tabular}
		\label{fig:depth95}
	}
	\vfil
	\subfloat[JPEG compression level: 15]{
		\begin{tabular}{c}
			\includegraphics[width=0.95\linewidth]{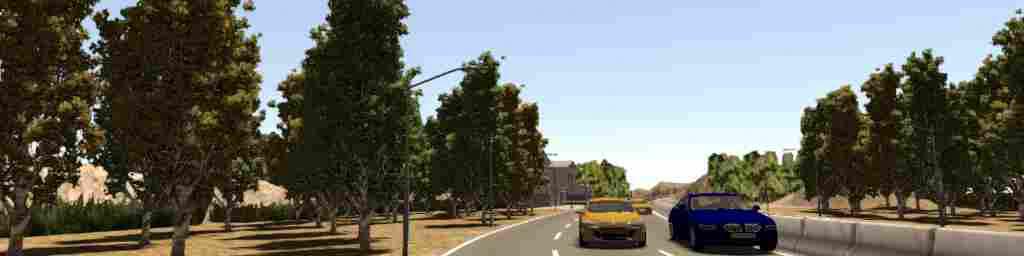}\\
			\includegraphics[width=0.95\linewidth]{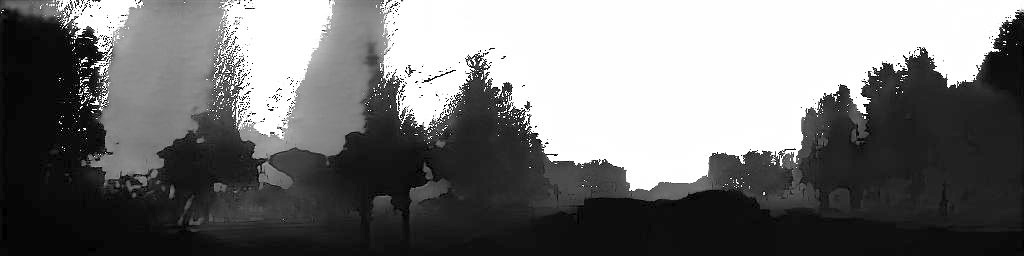}
		\end{tabular}
		\label{fig:depth15}
	}
	\vfil
	\subfloat[JPEG compression level: 10]{
		\begin{tabular}{c}
			\includegraphics[width=0.95\linewidth]{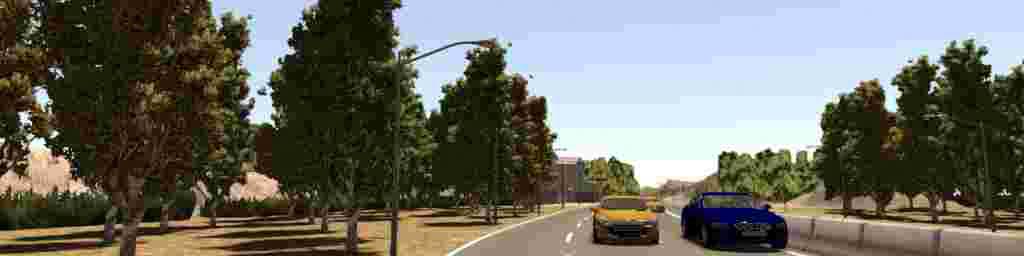}\\
			\includegraphics[width=0.95\linewidth]{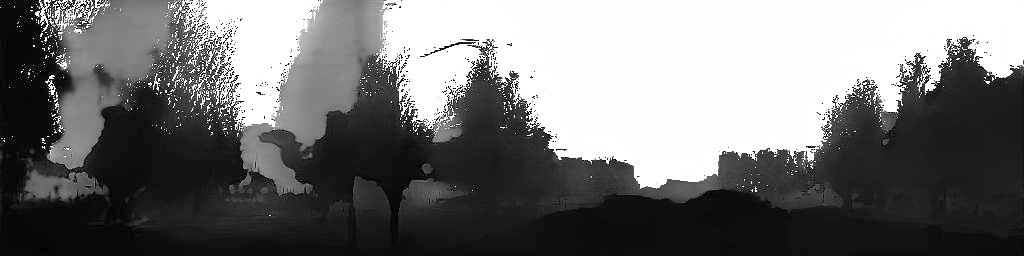}
		\end{tabular}
		\label{fig:depth10}
	}
	\setcounter{figure}{2}
\end{figure}

\subsection{Object Detection} \label{sec:fasterrcnn}
In object detection, we must locate and classify foreground objects within a scene (as opposed to semantic segmentation, which classifies each pixel), and compute the confidence of each classification (Fig. \ref{fig:fasterrcnn95}). We evaluate mAP of the Detectron FasterRCNN \cite{FasterRCNN} implementation \cite{Detectron} against the Pascal VOC 2007 dataset \cite{VOCdataset}, over mIoU with threshold 0.5:0.95. When training the network, we use $\eta=0.001$ and weight decay of 0.0005 over 60000 epochs.

\begin{figure}[!t]
	\vspace{-1cm}
	\centering
	\caption{Results of pre-trained FasterRCNN model \cite{FasterRCNN} on a JPEG image under different compression levels}
	\label{fig:fasterrcnn}
	\setcounter{figure}{2}
	\subfloat[JPEG compression level: 95]{\includegraphics[trim={1cm 1cm 1cm 1cm},clip, width=\linewidth]{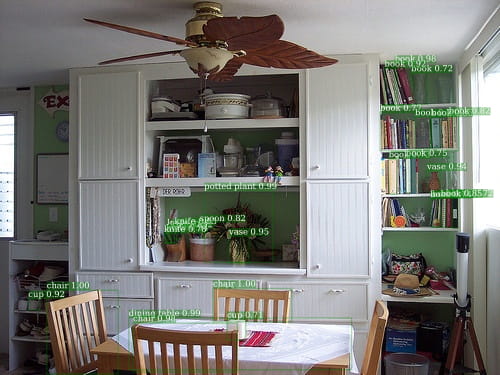}
	\label{fig:fasterrcnn95}}
	\vfil
	\subfloat[JPEG compression level: 15]{\includegraphics[trim={1cm 1cm 1cm 1cm},clip, width=\linewidth]{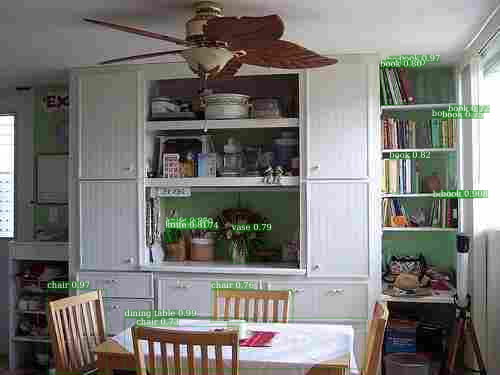}
	\label{fig:fasterrcnn15}}
	\vfil
	\subfloat[JPEG compression level: 10]{\includegraphics[trim={1cm 1cm 1cm 1cm},clip, width=\linewidth]{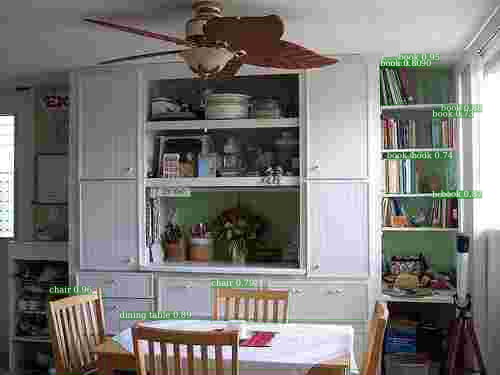}
	\label{fig:fasterrcnn10}}
	\setcounter{figure}{3}
\end{figure}

\subsection{Human Pose Estimation} \label{sec:pose}
Human Pose Estimation involves computing (and overlaying) the skeletal position of people detected within a scene (Fig. \ref{fig:openpose95}). Recent work uses part affinity fields to map body parts to individuals, thus distinguishing between visually similar features.

Using OpenPose \cite{OpenPose} we compute the skeletal overlay of detected people in images from the COCO dataset \cite{COCOdataset}. We evaluate with mean average precision (mAP), over 10 object key-point similarity (OKS) thresholds, where OKS represents IoU scaled over person size. When retraining the network, we use $\eta=0.001$, and a batch size of 8 over 40 epochs.

\begin{figure}[!t]
	\centering
	\caption{Results of pre-trained OpenPose model \cite{OpenPose} on a JPEG image under different compression levels}
	\label{fig:openpose}
	\setcounter{figure}{3}
	\subfloat[JPEG compression level: 95]{\includegraphics[trim={1cm 1cm 1cm 1cm},clip, width=\linewidth]{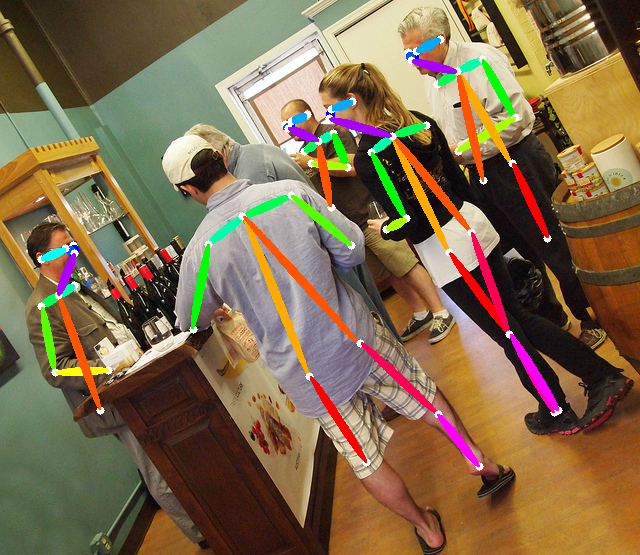}
	\label{fig:openpose95}}
	\vfil
	\subfloat[JPEG compression level: 15]{\includegraphics[trim={1cm 1cm 1cm 1cm},clip, width=\linewidth]{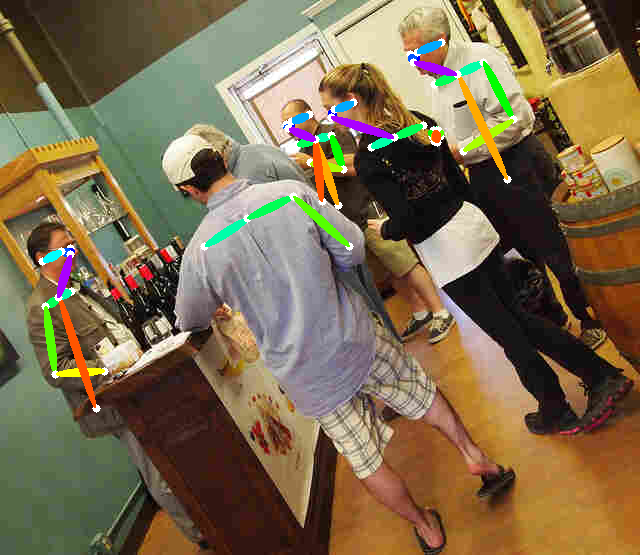}
	\label{fig:openpose15}}
	\vfil
	\subfloat[JPEG compression level: 10]{\includegraphics[trim={1cm 1cm 1cm 1cm},clip, width=\linewidth]{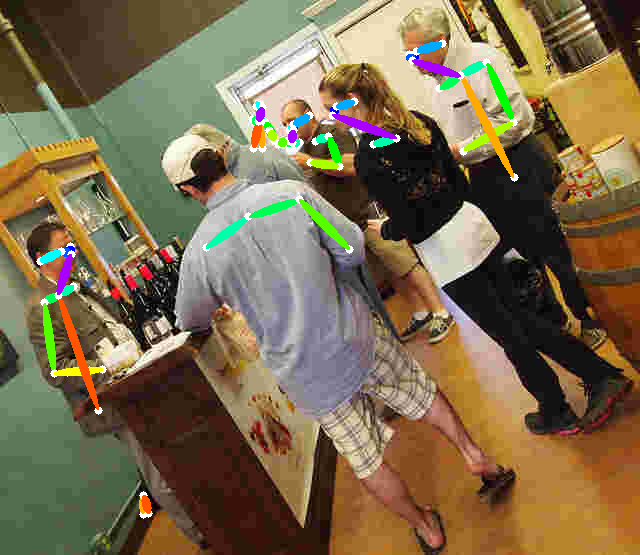}
	\label{fig:openpose10}}
	\setcounter{figure}{4}
\end{figure}

\subsection{Human Action Recognition} \label{sec:har}
To classify a single human action - from a handstand to knitting - with a reasonable level of accuracy, we must inspect spatial information from each frame, and temporal information across the entire video sequence.

We implement the dual-stream model from \cite{HAR}; recognising human activity by fusing spatial and temporal predictions from the UCF101 video dataset presented in \cite{UCF101} (see Fig. \ref{fig:har} for example frames, dramatically deteriorating in quality as H.264 CRF value is increased). To train the temporal stream, we pass 20 frames randomly sampled from the pre-computed stack of optical flow images. Across both streams, we use a batch size of 12, $\beta=0.9$, and  $\eta=0.001$ for 500 epochs.

\begin{figure}[!t]
	\centering
	\caption{One frame taken from a video input to the Two-Stream CNN model \cite{HAR} under different H.264 compression rates}
	\label{fig:har}
	\setcounter{figure}{4}
	\subfloat[H.264 CRF value 23]{\includegraphics[trim={1cm 1cm 1cm 1cm},clip, width=\linewidth]{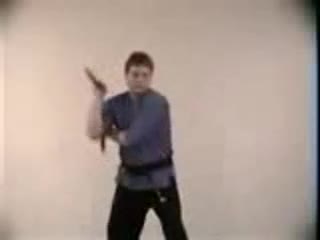}
		\label{fig:har23}}
	\vfil
	\subfloat[H.264 CRF value 30]{\includegraphics[trim={1cm 1cm 1cm 1cm},clip, width=\linewidth]{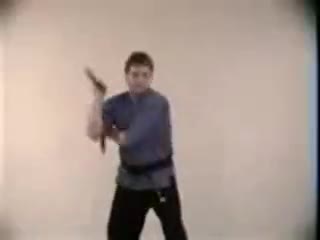}
		\label{fig:har30}}
	\vfil
	\subfloat[H.264 CRF value 40]{\includegraphics[trim={1cm 1cm 1cm 1cm},clip, width=\linewidth]{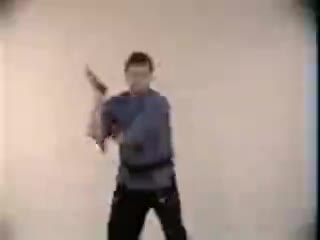}
	\label{fig:har40}}
	\setcounter{figure}{5}
\end{figure}

\begin{table}[!t]
	\caption{\textbf{segmentation:} Global accuracy, mean class accuracy and mIoU at varying compression rates}
	\label{tab:segment}
	\setcounter{table}{0}
	\centering
	\captionsetup[subtable]{justification=centering}
	\subfloat[after testing a pre-trained SegNet model \cite{SegNet} on compressed imagery]{
		\begin{tabular}{|c|c|c|c|}
			\hline
			Compression Rate & global ACC & mean ACC & mIoU \\
			\hline
			95 & 0.911 & 0.536 & 0.454\\
			\hline
			75 & 0.909 & 0.530 & 0.448\\
			\hline
			50 & 0.904 & 0.523 & 0.438\\
			\hline
			15 & 0.814 & 0.459 & 0.338\\
			\hline
			10 & 0.794 & 0.421 & 0.304\\
			\hline
			5 & 0.782 & 0.364 & 0.265\\
			\hline
		\end{tabular}
	\label{tab:segmentpre}}

	\captionsetup[subtable]{justification=centering}
	\subfloat[after retraining a SegNet model \cite{SegNet} with compressed imagery]{
		\begin{tabular}{|c|c|c|c|}
			\hline
			Compression Rate & global ACC & mean ACC & mIoU \\
			\hline
			95 & 0.911 & 0.536 & 0.454\\
			\hline
			75 & 0.910 & 0.522 & 0.446 \\
			\hline
			50 & 0.908 & 0.503 & 0.431 \\
			\hline
			15 & 0.902 & 0.494 & 0.420 \\
			\hline
			10 & 0.895 & 0.477 & 0.405 \\
			\hline
			5 & 0.879 & 0.445 & 0.374 \\
			\hline
		\end{tabular}
	\label{tab:segmentre}}
	\setcounter{table}{1}
\end{table}

\begin{table}[!t]
\caption{\textbf{Depth Estimation:} Absolute Relative,  Squared Relative, and Root Mean Squared Error at varying compression rates (lower, better)}
\label{tab:depth}
\setcounter{table}{1}
\centering
\captionsetup[subtable]{justification=centering}
\subfloat[after testing a pre-trained GAN model for monocular depth estimation \cite{Depth} on compressed imagery]{\begin{tabular}{|c|c|c|c|}
		\hline
		Compression Rate & Abs. Rel. & Sq. Rel. & RMSE \\
		\hline
		95 & 0.0112 & 0.0039 & 0.0588 \\
		\hline
		75 & 0.0116 & 0.0039 & 0.0589 \\
		\hline
		50 & 0.0123 & 0.0038 & 0.0587 \\
		\hline
		15 & 0.0146 & 0.0040 & 0.0599 \\
		\hline
		10 & 0.0192 & 0.0042 & 0.0617 \\
		\hline
		5 & 0.0283 & 0.0060 & 0.0749 \\
		\hline
	\end{tabular}
	\label{tab:depthpre}}

\captionsetup[subtable]{justification=centering}
\subfloat[retraining a GAN model for monocular depth estimation \cite{Depth} with compressed imagery]{\begin{tabular}{|c|c|c|c|}
		\hline
		Compression Rate & Abs. Rel. & Sq. Rel. & RMSE \\
		\hline
		95 & 0.0112 & 0.0039 & 0.0588 \\
		\hline
		75 & 0.0113 & 0.0035 & 0.0560 \\
		\hline
		50 & 0.0103 & 0.0029 & 0.0502 \\
		\hline
		15 & 0.0121 & 0.0034 & 0.0556 \\
		\hline
		10 & 0.0152 & 0.0031 & 0.0528 \\
		\hline
		5 & 0.0159 & 0.0040 & 0.0599 \\
		\hline
	\end{tabular}
	\label{tab:depthre}}
\setcounter{table}{2}
\end{table}

\begin{table}[!t]
	\caption{\textbf{object detection:}	Mean average precision at varying compression rates}
	\label{tab:fasterRCNN}
	\setcounter{table}{2}
	\centering
	\captionsetup[subtable]{justification=centering}
	\subfloat[after testing a pre-trained FasterRCNN model \cite{FasterRCNN} on compressed imagery]{\begin{tabular}{|c|c|}
			\hline
			Compression Rate & mAP \\
			\hline
			95 & 0.703 \\
			\hline
			75 & 0.686 \\
			\hline
			50 & 0.666 \\
			\hline
			15 & 0.545 \\
			\hline
			10 & 0.442 \\
			\hline
			5 & 0.187 \\
			\hline
		\end{tabular}
		\label{tab:fasterrcnnpre}}
	\captionsetup[subtable]{justification=centering}
	\subfloat[retraining a FasterRCNN model \cite{FasterRCNN} with compressed imagery]{\begin{tabular}{|c|c|}
			\hline
			Compression Rate & mAP \\
			\hline
			95 & 0.703 \\
			\hline
			75 & 0.694 \\
			\hline
			50 & 0.692 \\
			\hline
			15 & 0.647 \\
			\hline
			10 & 0.627 \\
			\hline
			5 & 0.559 \\
			\hline
		\end{tabular}
		\label{tab:fasterrcnnre}}
	\setcounter{table}{3}
\end{table}
\clearpage
\section{Evaluation}

In this section, we contrast the performance of the considered CNN architectures under their respective evaluation metrics before and after retraining. From this, we can determine how much we can safely compress the imagery while maintaining acceptable performance. We then propose possible explanations for the variations in resilience of the network architectures to image compression.

\subsection{Semantic Segmentation} \label{sec:segmentation2}
From results presented in Table \ref{tab:segment} we can observe that the impact of lossy compression (Table \ref{tab:segmentpre}) is minimal, indicating high resilience to compression within the network. At the highest (most compressed) compression level, we see global accuracy reduce by 14\%, down to 78.2\%, while affording 95\% less storage cost on average per input image. However, at these heaviest compression rates, the compression artifacts introduced can lead to false labelling. This is particularly prominent where there are varying levels of lighting, affecting even plain roads (Fig. \ref{fig:segm10}). Subsequently, from Table \ref{tab:segmentre} we can see that retraining the network further minimizes performance loss, especially minimizing false labelling of regions. At a JPEG compression level of 5, performance loss is reduced to 3.5\%, resulting in global accuracy narrowly dropping below 0.9. Such resilience may stem from the up-sampling by the pooling layers within the decoder pipeline, which are innately capable of recovering information that has been lost during compression, but further investigation is left to future work.

\subsection{Depth Estimation}
Analyzing the results in Table \ref{tab:depth}, it is evident that lossy compression markedly diminishes RMSE performance of depth estimation when heavy compression rates are employed (Table \ref{tab:depthpre}). At a JPEG compression level of 15, RMSE has not increased by more than 1.9\%, but at a JPEG compression level of 10 and lower, performance begins to dramatically decline (in keeping with that of \cite{AVG}). However, by retraining the network at the same compression level that is employed during testing (Table \ref{tab:depthre}), performance loss can be thoroughly constrained. Even at a JPEG compression level of 5, RMSE can be constrained to under 0.0600, improving performance by as much as 20\% over the pre-trained network. Other performance measures demonstrate the same trend.

This performance is surprising: we might expect that RMSE would increase (thus lowering performance) after training on compressed imagery, since the GAN generates low quality imagery as the textures and features used to calculate depth estimation are lost, and is therefore unable to improve depth estimation performance. It is possible that it exceeds our expectation due to the encoder-decoder pipeline within the estimation process, which is also employed in the SegNet architecture, and thereby shares its compression resilience.

\subsection{Object Detection}

From Table \ref{tab:fasterRCNN}, we can again discern that performance degrades rapidly at high lossy compression levels (JPEG compression level of 15 or less, see Table \ref{tab:fasterrcnnpre}). Applying a JPEG compression level of 15 leads to a 22.5\% drop, down to mAP of 0.545, while a JPEG compression level of 5 causes mAP to drop by as much as 73.4\%. Furthermore, with higher compression rates, fewer objects are detected, and their classification confidence also falls (Fig. \ref{fig:fasterrcnn10}). Their classification accuracy remains unhindered, however. When the network is retrained on imagery lossily compressed at the same level, performance is noticeably improved (Table \ref{tab:fasterrcnnre}). The performance drop as compression rate is increased is delayed from a JPEG compression level of 15 to a JPEG compression level of 5. In fact, the retrained network is able to maintain an mAP above 0.6 even at a JPEG compression level of 10; reducing performance degradation to only 10.8\%, while affording a lossy compression rate almost 10-fold higher in terms of reduced image storage requirements. 

\subsection{Human Pose Estimation}

\begin{table}[!t]
	\caption{\textbf{human pose estimation:} Mean average precision at varying compression rates}
	\label{tab:pose}
	\setcounter{table}{3}
	\centering
	\captionsetup[subtable]{justification=centering}
	\subfloat[after testing a pre-trained OpenPose model \cite{OpenPose} on compressed imagery]{\begin{tabular}{|c|c|}
			\hline
			Compression Rate & mAP \\
			\hline
			95 & 0.711 \\
			\hline
			75 & 0.689 \\
			\hline
			50 & 0.655 \\
			\hline
			15 & 0.413 \\
			\hline
			10 & 0.323 \\
			\hline
			5 & 0.098 \\
			\hline
		\end{tabular}
	\label{tab:posepre}}
	\captionsetup[subtable]{justification=centering}
	\subfloat[after retraining an OpenPose model \cite{OpenPose} with compressed imagery]{\begin{tabular}{|c|c|}
			\hline
			Compression Rate & mAP \\
			\hline
			95 & 0.711 \\
			\hline
			75 & 0.708 \\
			\hline
			50 & 0.678 \\
			\hline
			15 & 0.654 \\
			\hline
			10 & 0.597 \\
			\hline
			5 & 0.454 \\
			\hline
		\end{tabular}
	\label{tab:posere}}
	\setcounter{table}{4}
\end{table}
Results in Table \ref{tab:pose} once again illustrate that lossy image compression (Table \ref{tab:posepre}) dramatically impacts performance at high rates. Similar to object detection, performance considerably lowers at 15\% compression rate, in this case with performance falling by 41.9\% to 0.413 mAP. Qualitatively, the network computes precisely located skeletal positions at higher compression rates, but detects and locates fewer joints (Fig. \ref{fig:openpose15}). With high levels of compression (Fig. \ref{fig:openpose10}), the false positive rate increases, and limbs are falsely detected and located. It is likely that optimizing the detection confidence threshold required of joints before computing their location, and thereby maximizing limb detection while minimizing false positives increases performance, especially during high compression. With a retrained network (Table \ref{tab:posere}), a compression rate of 15\% can be safely achieved before performance degradation exceeds 10\%.

While impressive, the results are relatively insubstantial compared to those of other architectures, such as SegNet (Section \ref{sec:segmentation2}, Table \ref{tab:segment}). The difference can perhaps be attributed to the double prediction task within the pose estimation network. Inaccuracies stemming from the lower quality images are not just propagated but multiplied through the network, as the architecture must simultaneously predict both detection confidence maps and the affinity fields for association encodings.

\subsection{Human Action Recognition}

\begin{table}[!t]
	\caption{\textbf{human action recognition:} Top-1 accuracy for each stream at varying compression rates}
	\label{tab:HAR}
	\setcounter{table}{4}
	\centering
	\captionsetup[subtable]{justification=centering}
	\subfloat[after testing a pre-trained HAR model \cite{HAR} on video data with varying H.264 CRF encoding values]{\begin{tabular}{|c|c|c|c|}
			\hline
			Compression Rate & Top-1 Spatial & Top-1 Motion & Top-1 Fusion\\
			\hline
			23 & 78.8736 & 70.1198 & 83.5485 \\
			\hline
			25 & 78.7999 & 44.9225 & 73.6030 \\
			\hline
			30 & 78.4563 & 37.3598 & 72.2329 \\
			\hline
			40 & 74.5704 & 38.9565 & 70.8803 \\
			\hline
			50 & 44.1977 & 15.3267 & 41.4777 \\
			\hline
		\end{tabular}
		\label{tab:harpre}}
	
	\captionsetup[subtable]{justification=centering}
	\subfloat[after retraining a HAR model \cite{HAR} with on video data with varying H.264 CRF encoding values]{\begin{tabular}{|c|c|c|c|}
			\hline
			Compression Rate & Top-1 Spatial & Top-1 Motion & Top-1 Fusion\\
			\hline
			23 & 78.8736 & 70.1198 & 83.5485 \\
			\hline
			25 & 78.9056 & 39.7192 & 71.7616\\
			\hline
			30 & 78.5620 & 34.3161 & 70.5765\\
			\hline
			40 & 75.9450 & 9.2550 & 67.1227\\
			\hline
			50 & 62.5165 & 6.7300 & 56.2279\\
			\hline
		\end{tabular}
		\label{tab:harre}}
	\setcounter{table}{5}
\end{table}
From results presented in Table \ref{tab:HAR}, it is evident that the impact of lossy compression (Table \ref{tab:harpre}) dramatically increases when we apply CRF factor 50. Conversely to all other examined architectures, we can see from Table \ref{tab:harre} that retraining the network in fact \textit{decreases} performance.

At first glance, we might expect similar performance to pose detection as with the two stream network for human action recognition, as the errors introduced by compression artifacts propagate through both streams in the network. However, the spatial and motion streams are not trained in tandem. While the spatial stream remains resilient, once again due to the up-sampling within the architecture (Section \ref{sec:segmentation2}), the motion stream is almost entirely unable to learn from compressed imagery. As such, retraining the network on compressed imagery in fact reduces overall performance (aside from when using CRF 50, as the spatial stream improvement outweighs the motion stream degradation). Future work may reveal whether better performance might be achieved by retraining just the spatial stream network on compressed imagery, and fusing its predictions with a motion stream trained only on uncompressed imagery. 

%%%%%%%%%%%%%%%%%%%%%%%%DISCUSSION%%%%%%%%%%%%%%%%%%%%%%%%%%%%%%%%
%%%%%%%%%%%%%%%%%%%%%%%%%%%%%%%%%%%%%%%%%%%%%%%%%%%%%%%%%%%%%%%%%%%%
\section{Conclusion}
This study has investigated the impact of lossy image compression on a multitude of existing deep CNN architectures. We have considered how much compression can be achieved while maintaining acceptable performance, and to what extent performance degradation can be ameliorated by retraining the networks with compressed imagery.

Across all challenges, retraining the network on compressed imagery recovers performance to a certain degree. This study has brought to attention in particular, however, that in very prevalent and so far unexamined network architectures, we can afford to compress imagery at extremely high rates. Segmentation and depth estimation in particular demonstrate resilience against even very significant compression, both by employing an encoder-decoder pipeline. By using retrained models, compression can safely reach as high as 85\% across all domains. In doing so, current storage costs can be markedly diminished before performance is noticeably impacted. Hyper parameter optimization of the retrained model can assumedly capitalize on this even further, and in certain domains, such as segmentation, we can already afford to reduce to a twentieth of the original storage cost. It should be noted however, that even a 1 or 2\% performance loss may be unacceptable in safety critical operations, such as depth estimation for vehicular visual odometry.

We can further suggest that lossy image compression is potentially viable as a data augmentation technique within RCNN \cite{FasterRCNN} and pose estimation \cite{OpenPose} architectures, which receive only mild performance degradation. Networks employing an encoder-decoder architecture (SegNet \cite{SegNet}, GAN \cite{Depth}) would only notably benefit from very significant levels of image compression for data augmentation. However, human action recognition networks, or sub-networks in the case of the two stream approach \cite{HAR}, that consider motion input will not readily benefit from image compression as a data augmentation technique, since they appear unable to learn under such training conditions.

%Future work should investigate optimal data splits, biases and other techniques to address the problem of unbalanced-classes within training data, when adopting different compression levels for training networks, which might influence network weights during training. 
Future work will investigate whether performance is improved by retraining the network with more heavily or lightly compressed imagery than at testing, or even a variety of compression levels. Furthermore, evaluating performance of compressed networks such as MobileNet\cite{mobilenet} against compressed imagery would be pertinent, as such light network architectures are prevalent amidst compressed imagery application domains.

%%%%%%%%%%%%%%%%%%%%%%%%ACKNOWLEDGEMENT%%%%%%%%%%%%%%%%%%%%%%%%%%%%%%%%
%%%%%%%%%%%%%%%%%%%%%%%%%%%%%%%%%%%%%%%%%%%%%%%%%%%%%%%%%%%%%%%%%%%%
\section{Acknowledgements}
This work was supported by Durham University, the European Regional Development Fund Intensive Industrial Innovation Grant No. 25R17P01847
\bibliographystyle{./IEEEtran}
\bibliography{./compression}
%\begin{thebibliography}{1}
%\end{thebibliography}

\end{document}